\documentclass{article}

\PassOptionsToPackage{numbers, compress}{natbib}
\usepackage[final]{nips_2017}

\usepackage{microtype,marvosym}

\usepackage{tikz,pgfplots}
\pgfplotsset{compat=newest}
\pgfplotsset{plot coordinates/math parser=false}

\usetikzlibrary{arrows,shapes,plotmarks,pgfplots.colormaps}
\tikzset{>=stealth'} 
\tikzstyle{graphnode} = 
   [circle,draw=black,minimum size=22pt,text centered,text
     width=22pt,inner sep=0pt] 
\tikzstyle{var}   =[graphnode,fill=white]
\tikzstyle{obs}   =[graphnode,fill=black,text=white]
\tikzstyle{fac}   =[rectangle,draw=black,fill=black!25,minimum size=5pt]
\tikzstyle{facprior} =[rectangle,draw=black,fill=black,text=white,minimum size=5pt]
\tikzstyle{edge}  =[draw=white,double=black,thick,-]
\tikzstyle{prior} =[rectangle, draw=black, fill=black, minimum size=
5pt, inner sep=0pt]
\tikzstyle{dirprior} = [circle, draw=black, fill=black, minimum
size=5pt, inner sep=0pt]

\pgfplotsset{compat=newest}
\pgfplotsset{
  every axis legend/.append style =
    {
      cells = { anchor = east },
      draw  = none
    },
}  
\pgfkeys{/pgfplots/mystyle/.style={
  semithick,
  tick style={major tick length=4pt,semithick,gray},
  xtick align = inside,
  ytick align = inside
  }}

\newcommand{\g}{\,|\,}

\newcommand{\Exp}{\mathbb{E}}

\renewcommand{\Re}{\mathbb{R}}

\newcommand{\Trans}{^{\intercal}} 
\newcommand{\argmin}{\operatorname*{arg\:min}}

\newcommand{\q}{\quad}
\newcommand{\qq}{\qquad}

\renewcommand{\vec}{\boldsymbol}

\newcommand{\GP}{\mathcal{GP}}

\usepackage{colonequals}
\newcommand{\ce}{\colonequals}

\DeclareSymbolFont{stmry}{U}{stmry}{m}{n}
\DeclareMathSymbol\leftarrowtriangle\mathrel{stmry}{"5E}
\DeclareMathSymbol\rightarrowtriangle\mathrel{stmry}{"5F}
\DeclareMathSymbol\leftrightarrowtriangle\mathrel{stmry}{"5D}
\DeclareMathSymbol\obar\mathrel{stmry}{"3A}
\DeclareMathSymbol\otimes\mathrel{stmry}{"0F}
\DeclareMathSymbol\ominus\mathrel{stmry}{"17}
\DeclareMathSymbol\sslash\mathrel{stmry}{"0C}

\usepackage{mathtools}
\usepackage{algorithm}
\usepackage{algpseudocode}
\algrenewcommand{\algorithmiccomment}[1]{\hfill {\footnotesize $\sslash$ #1}}
\algrenewcommand{\alglinenumber}[1]{\tt\scriptsize #1}

\makeatletter
\@addtoreset{algorithm}{chapter} 
\makeatother

\makeatletter
\def\therule{\makebox[\algorithmicindent][l]{\hspace*{.5em}\vrule height .75\baselineskip depth .25\baselineskip}}%

\newtoks\therules
\therules={}
\def\appendto#1#2{\expandafter#1\expandafter{\the#1#2}}
\def\gobblefirst#1{
  #1\expandafter\expandafter\expandafter{\expandafter\@gobble\the#1}}%
\def\LState{\State\unskip\the\therules}
\def\pushindent{\appendto\therules\therule}%
\def\popindent{\gobblefirst\therules}%
\def\printindent{\unskip\the\therules}%
\def\printandpush{\printindent\pushindent}%
\def\popandprint{\popindent\printindent}%

\algdef{SE}[WHILE]{While}{EndWhile}[1]
  {\printandpush\algorithmicwhile\ #1\ \algorithmicdo}
  {\popandprint\algorithmicend\ \algorithmicwhile}%
\algdef{SE}[FOR]{For}{EndFor}[1]
  {\printandpush\algorithmicfor\ #1\ \algorithmicdo}
  {\popandprint\algorithmicend\ \algorithmicfor}%
\algdef{S}[FOR]{ForAll}[1]
  {\printindent\algorithmicforall\ #1\ \algorithmicdo}%
\algdef{SE}[LOOP]{Loop}{EndLoop}
  {\printandpush\algorithmicloop}
  {\popandprint\algorithmicend\ \algorithmicloop}%
\algdef{SE}[REPEAT]{Repeat}{Until}
  {\printandpush\algorithmicrepeat}[1]
  {\popandprint\algorithmicuntil\ #1}%
\algdef{SE}[IF]{If}{EndIf}[1]
  {\printandpush\algorithmicif\ #1\ \algorithmicthen}
  {\popandprint\algorithmicend\ \algorithmicif}%
\algdef{C}[IF]{IF}{ElsIf}[1]
  {\popandprint\pushindent\algorithmicelse\ \algorithmicif\ #1\ \algorithmicthen}%
\algdef{Ce}[ELSE]{IF}{Else}{EndIf}
  {\popandprint\pushindent\algorithmicelse}%
\algdef{SE}[PROCEDURE]{Procedure}{EndProcedure}[2]
   {\printandpush\algorithmicprocedure\ \textproc{#1}\ifthenelse{\equal{#2}{}}{}{(#2)}}%
   {\popandprint\algorithmicend\ \algorithmicprocedure}%
\algdef{SE}[FUNCTION]{Function}{EndFunction}[2]
   {\printandpush\algorithmicfunction\ \textproc{#1}\ifthenelse{\equal{#2}{}}{}{(#2)}}%
   {\popandprint\algorithmicend\ \algorithmicfunction}%
\makeatother

\usetikzlibrary{external}
\tikzset{external/force remake=false}
\tikzsetexternalprefix{fig/external/}

\graphicspath{{fig/}}

\newlength{\figheight}
\newlength{\figwidth}

\usepackage[utf8]{inputenc} 
\usepackage[T1]{fontenc}    
\usepackage[colorlinks   = true, 
  urlcolor     = blue, 
  linkcolor    = blue, 
  citecolor   = red 
  ]{hyperref}       
\usepackage{url}            
\usepackage{booktabs}       
\usepackage{amsfonts,amsmath,amssymb,mathtools}       
\usepackage{nicefrac}       
\usepackage{microtype}      

\newcommand{\spa}{\operatorname{span}}

\title{Krylov Subspace Recycling for\\{Fast} Iterative Least-Squares in Machine Learning}

\author{
  Filip de Roos and Philipp Hennig\\
  Max Planck Institute for Intelligent Systems
  Spemannstr. 34, T\"ubingen, Germany\\
  \texttt{[fderoos|ph]@tue.mpg.de}
}

\begin{document}
\maketitle

\begin{abstract}
  Solving symmetric positive definite linear problems is a fundamental computational task in machine learning. The exact solution, famously, is cubicly expensive in the size of the matrix. To alleviate this problem, several linear-time approximations, such as spectral and inducing-point methods, have been suggested and are now in wide use. These are low-rank approximations that choose the low-rank space a priori and do not refine it over time. While this allows linear cost in the data-set size, it also causes a finite, uncorrected approximation error. Authors from numerical linear algebra have explored ways to iteratively refine such low-rank approximations, at a cost of a small number of matrix-vector multiplications. This idea is particularly interesting in the many situations in machine learning where one has to solve a sequence of related symmetric positive definite linear problems. From the machine learning perspective, such \emph{deflation} methods can be interpreted as transfer learning of a low-rank approximation across a time-series of numerical tasks. We study the use of such methods for our field. Our empirical results show that, on regression and classification problems of intermediate size, this approach can interpolate between low computational cost and numerical precision.
\end{abstract}

\section{Introduction}
\label{sec:introduction}

Many of the most prominent machine learning problems can be seen as a sequence of linear systems, finding $\vec{x}^{(i)}$ such that 
\begin{equation}\label{eq:task}
A^{(i)} \vec{x}^{(i)} = \vec{b}^{(i)}\qq \text{for}\qq A^{(i)}\in\Re^{n\times n},\q \vec{x}^{(i)},\vec{b}^{(i)}\in\Re^{n}, \qq\text{and}\qq i\in\mathbb{N}.	
\end{equation}
A prominent example is nonparametric logistic regression, further explained below. But there are many more: Model adaptation in Gaussian process models \citep[][\textsection5.2]{RW-GPML} requires the solution of the problem $k_{\theta,XX}^{-1}\vec{y}$ for a sequence of parameter estimates $\theta$, where $k_{XX}$ is a kernel Gram matrix over the data set $X=[\vec{x}_1,\dots,\vec{x}_n]\Trans$, and $\vec{y}$ is the vector of target data. Further afield, although this view is not currently the popular standard, deep learning tasks have in the past been addressed by methods like Hessian-free optimization \cite{martens10}, which consist of such sequences. Importantly, in these machine learning examples, the matrix $A$ is usually symmetric positive definite (or, in the generally non-convex case of deep learning, is at least approximated by an spd matrix in the optimizer to ensure a descent step). This means that Eq.~\eqref{eq:task} is indeed an optimization problem, because $\vec{x}^{(i)}$ then equals the minimum of the quadratic function 
\begin{equation}\label{eq:optimization-form}
	\vec{x}^{(i)} = \argmin_{\tilde{\vec{x}}} \frac{1}{2} \tilde{\vec{x}}\Trans A^{(i)} \tilde{\vec{x}} - \tilde{\vec{x}}\Trans \vec{b}^{(i)} \q\text{with gradient}\q \vec{r}^{(i)}(\tilde{\vec{x}}) \ce \nabla f(\tilde{\vec{x}}) = A^{(i)} \tilde{\vec{x}} - \vec{b}^{(i)}.
\end{equation}

When facing linear problems of small to moderate size (i.e.~$n\lesssim 10^4$) in machine learning , the typical approach is to rely on standard algorithms (in particular, Cholesky decompositions) provided by toolboxes like BLAS libraries, or iterative linear solvers like the method of conjugate gradients~\cite{hestenes52} (CG). Exact methods like the Cholesky decomposition have cubic cost, $\mathcal{O}(n^3)$, iterative solvers like CG have quadratic cost, $\mathcal{O}(n^2m)$ for a small number of $m$ iterative steps. These algorithms are self-contained, generic ``black boxes''---they are designed to work on \emph{any} spd matrix and approach every new problem in the same way. A more critical way to phrase this property is that these algorithms are \emph{non-adaptive}. If the sequence of tasks ($A^{(i)},\vec{b}^{(i)}$) are related to each other, for example because they are created by an outer optimization loop, it seems natural to want to propagate information from one linear solver to another. One can think of this notion as a form of \emph{computational transfer learning}, in the ``probabilistic numerics'' sense of treating a computation as an inference procedure~\cite{hennig2015probabilistic}. As it turns out, the computational linear algebra community has already addressed this issue to some extent. In that community, the idea of re-using information from previous problems in subsequent ones is known as \emph{subspace-recycling}~\cite{parks06}. But these numerical algorithms have not yet found their way into the machine learning community. Below, we explore the utility of such a resulting method for application in machine learning, by empirically evaluating it on the test problem of Bayesian logistic regression (aka. Gaussian process classification, GPC).

\subsection{Relation to Linear-Cost Methods}
\label{sub:relation_to_linear_cost_methods}

For linear problems of \emph{large} dimensionality (data-sets of size $n\gtrsim 10^4$), the current standard approach is to introduce an ``a-priori'' low-rank approximation: Sampling a small set of approximate eigenvectors of the kernel gram matrix~\cite{NIPS2007_3182,NIPS2008_3495}, and introducing various conditional independence assumptions over sets of inducing points~\cite{NIPS2000_1866,quinonero05,snelson07a}. These methods achieve \emph{linear} cost $\mathcal{O}(nm^2)$ because they project the problem onto a projective space of dimensionality $m$, and this space is not adapted over time (when it is adapted \cite{titsias09a}, additional computational overhead is created outside the solver). The downside of this approach is that it yields an approximation of finite quality---these methods fundamentally can not converge, in general, to the exact solution. The algorithms we consider below can be seen in some sense as the ``missing link'' between these linear-cost-but-finite-error methods and the cubic-cost, exact solvers for smaller problems: They \emph{adapt} the projective sub-space over time at the cost of a small number of quadratic cost steps, while also attempting to re-use as much information from previous runs as possible. In fact, the ``guessed'' projective space of the aforementioned methods could be used as the first initialization of the methods discussed below.

\section{Method}
\label{sec:method}

\begin{figure}
\centering
\def\svgwidth{\linewidth}
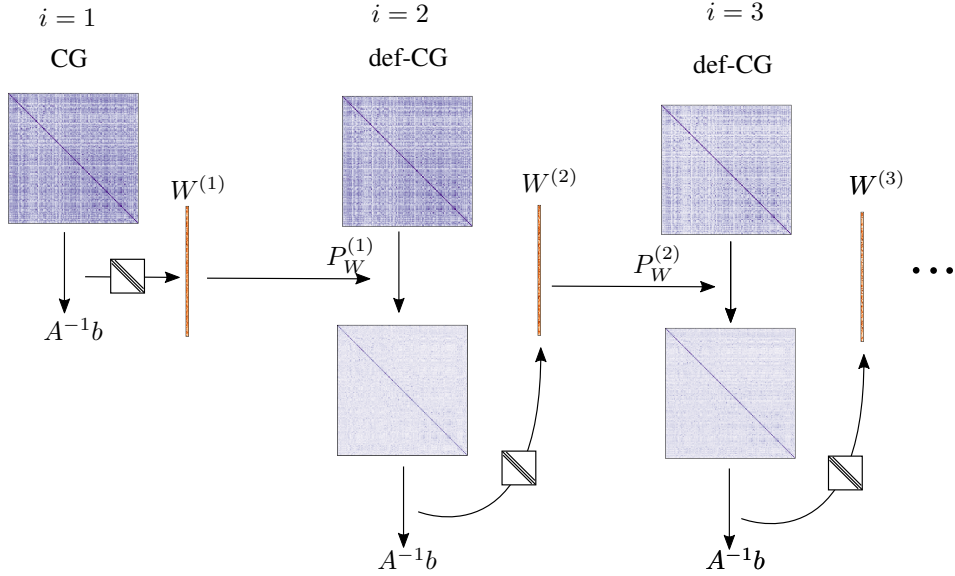
\caption{\label{fig:def_cg} The def-CG algorithm applied to a sequence of linear systems with the implicit preconditioning visualized. The first solution is obtained through normal CG and approximate eigenvectors $W$, corresponding to the largest eigenvalues are calculated from a low-rank approximation. If prior knowledge about the eigenvectors of the first system is available, def-CG can be used for the first system as well.}
\end{figure}

\subsection{Krylov Subspace Recycling}
The \emph{Krylov sequence} of subspaces is a core concept of iterative methods for linear systems $A\vec{x}=\vec{b}$~\cite{saad11}. The \emph{Krylov subspace} $\mathcal{K}_j(A,\vec{r}_0)$ is the span of the truncated power iteration of $A$ operating on $\vec{r}_0$, where $\vec{r}_{0}$ is the gradient (\emph{residual}) from Eq.~\eqref{eq:optimization-form} at the starting point $\vec{x}_0$ of the optimization:
\begin{equation}
\mathcal{K}_j(A,\vec{r}_0)=\spa\{\vec{r}_0,A\vec{r}_0,...,A^{j-1}\vec{r}_0\}.
\end{equation}
Methods that \emph{transfer} information from one such iteration to the next in order to faster converge to a solution in subsequent systems are referred to as \emph{Krylov subspace recycling methods}~\cite{parks06}. The ``recycling'' of information is traditionally done by \textit{deflation}~\cite{saad00,frank01}, \textit{augmentation}~\cite{gaul13,morgan95}, or combinations thereof~\cite{ebadi16,chapman97}. These two approaches differ in their implementation, but have the same goal: restricting the solution to a simpler search space, to speed up convergence. Both store a set of $k$ linearly independent vectors $W^{(i)}\in\mathbb{R}^{n\times k}$. For problem $i$ in the above sequence of tasks, a subspace-recycling Krylov method computes solutions that satisfy
\begin{align}
\vec{x}_j ^{(i)}&\in \vec{x}_0+\mathcal{K}_j(A,\vec{r}_0) \cup \spa\{W^{(i)}\}, \\
\vec{r}_j ^{(i)}&= \vec{b}-A\vec{x}_j \perp \mathcal{K}_j(A,\vec{r}_0) \cup \spa\{W^{(i)}\}. \label{eq:res}
\end{align}
An augmented iterative solver keeps the vectors in $W$ and orthogonalizes the updated residuals $\vec{r}_j ^{(i)}$ against $W$. This method is easily included in methods that contain an explicit orthogonalization step, an example of which is the General Minimum Residual method (GMRES)~\cite{morgan95}.\\
For $A$ spd, one usually chooses CG as the iterative solver and deflation is easier incorporated~\cite{saad00}. A deflated method ``deflates'' a part of the spectrum of $A$ by projecting the solution onto the orthogonal complement of $W$. This can be viewed as a form of preconditioning\footnote{Not all authors agree: \citet{gaul13} argue that the projector $P_W$ should not be considered a preconditioner since its application removes a part of the spectrum of A while leaving the remainder untouched.} with a singular projector $P_W$, i.e.~solving $P^{(i)}_W A^{(i)
} \vec{x}^{(i)}=P^{(i)}_W\vec{b}^{(i)}$. This property is in contrast to normal preconditioning where the whole spectrum of $A$ is modified by a non-singular matrix.  
 In order to keep the additional computational overhead incurred by subspace recycling low, the dimension of $W$ should be low, and contain vectors that optimally speed up the convergence. For iterative linear solvers, the rate of convergence is directly proportional to the condition number 
\begin{equation}
\kappa (A)=\frac{\lambda_{n}(A)}{\lambda_{1}(A)},
\end{equation} 
where $\lambda_j$ refers to the eigenvalues of $A$ sorted in ascending order \cite{nocedal06}. Typically, the smallest eigenvalues are the limiting factors for convergence; hence $W^{(i)}$ should ideally contain the eigenvectors related to the $k$ \emph{smallest} eigenvalues---this yields an effective condition number of $\kappa_{\text{eff}} = \nicefrac{\lambda_{n}}{\lambda_{k+1}}$, which can drastically improve the convergence rate~\cite{saad11,ebadi16}. Of course the same improvement in the condition number can also be achieved by changing the \emph{largest} eigenvalue.

\subsection{Deflated Conjugate Gradient}
\label{sec:def_cg}

In the special case when $A$ is symmetric and positive definite (SPD), the conjugate gradient method (CG) \cite{hestenes52} is a popular choice to iteratively solve the system. As noted above, in machine learning, where linear tasks almost invariably arise in the form of least-squares problems, this is actually the typical setting. \citet{saad00} derived a deflated version of CG based on the Lanczos iteration (partly represented in Algorithm \ref{alg:defcg}): it implicitly forms a tri-diagonal low-rank approximation to a Hermitian matrix $A$, which translates to symmetric when $A$ is real-valued. By storing quantities that are readily available from the CG iterations, the low-rank approximation can be obtained without costly matrix-vector computations. 
The eigenvalues of the approximation are known as Ritz values, which tend to approximate the extreme ends of the spectrum of $A$~\cite{saad11}. The corresponding Ritz vectors are used to find good approximations of the eigenvectors that can be used for a deflated subspace $W$ to improve the condition number \cite[see][for more details]{saad00}. 
The algorithm will be referred to as def-CG($k,\ell$) where $\ell$ is the number of CG iterations from which information is stored in order to generate $k$ approximate eigenvectors. 
\\
For deflated CG, the orthogonality constraint of Eq.~\eqref{eq:res} is replaced by a constraint of \emph{conjugacy}, i.e.~$\vec{r}_i\Trans A\vec{r}_j=0$ for $i\neq j$.
The associated preconditioner $P_W=I-AW(W^T A W)^{-1}W^T$ projects the residual onto the $A$-conjugate complement of $W$. Figure \ref{fig:def_cg} illustrates the effect of applying $P_W$ to $A$. For this figure, $W$ was chosen by the def-CG algorithm according to the harmonic projection method~\cite{morgan95}, so the approximate eigenvectors correspond to the largest eigenvalues. The algorithm (Algorithm \ref{alg:defcg}) differs from the standard method of conjugate gradients only in line 11 and the initialization in line 3. How the eigenvectors are approximated is outlined in Section \ref{sub:approximate_eigenvectors}.
The additional inputs to the solver are a set of $k$ linearly independent vectors in $W$ and, optionally, $W$ multiplied with $A$ if it can be obtained cheaply. \\
To estimate the computational cost of def-CG we assume $k\ll n$ so terms not containing $n$ in the computational complexity can be ignored. Each iteration in Algorithm \ref{alg:defcg} has a computational overhead of $\mathcal{O}(kn^2)$ of solving the linear system in line 11. Computing the matrix $W^TAW$ has complexity $\mathcal{O}(n^2k^2)$ but it only has to be computed once and if the procedure used by \citet{saad00} and further outlined in section \ref{sub:approximate_eigenvectors}, $W$ and $AW$ are obtained in $\mathcal{O}(n^2(\ell+1)k)$. By choosing $\ell$ and $k$ to be small the computational overhead can be kept modest. This shows the importance of choosing vectors in $W$ that significantly reduce the number of required iterations to make up for the computational overhead. Another factor to take into account is the additional storage requirements of the deflated CG. The main contributing factors are the matrices $W$ and $AW$, which each are of size $n\times k$ and $\ell$ search directions of size $n$. 

\begin{algorithm}
\begin{algorithmic}[1]
\Procedure{Deflated-CG($k,\ell$)}{$A$, $b$, $x_{-1}$, $W$, $(AW)$, tol}
\Ensure{$W\in\Re^{n\times k}$} \Comment{$k$ included in Alg. definition for interpretability, can obviously be inferred internally.}
\LState \raisebox{0pt}[0pt][0pt]{Choose $x_0$ such that $W^Tr_0=0$ where $r_0=b-Ax_0$}
\LState \raisebox{0pt}[0pt][0pt]{$x_0=x_{-1}+W(W^TAW)^{-1}W^Tr_{-1}$}
\LState \raisebox{0pt}[0pt][0pt]{Solve $W^TAW\mu_0=W^TAr_0$ for $\mu$ and set $p_0=r_0-W\mu_0$} \Comment{deflation for initial iteration}
\While{\raisebox{0pt}[0pt][0pt]{$|r_j|>\text{tol}$ (For j=1...)}}
\LState \raisebox{0pt}[0pt][0pt]{$\phantom{\alpha_{j-1}}\mathllap{d_{j-1}}=p_{j-1}^TAp_{j-1}$} 
\LState \raisebox{0pt}[0pt][0pt]{$\phantom{\alpha_{j-1}}\mathllap{\alpha_{j-1}}=r_{j-1}^Tr_{j-1}/{d_{j-1}}$}
\LState \raisebox{0pt}[0pt][0pt]{$\phantom{\alpha_{j-1}}\mathllap{x_j}=x_{j-1}+\alpha_{j-1}p_{j-1}$}
\LState \raisebox{0pt}[0pt][0pt]{$\phantom{\alpha_{j-1}}\mathllap{r_j}=r_{j-1}-\alpha_{j-1}Ap_{j-1}$}
\LState \raisebox{0pt}[0pt][0pt]{$\phantom{\alpha_{j-1}}\mathllap{\beta_{j-1}}=r_{j}^Tr_{j}/r_{j-1}^Tr_{j-1}$} \Comment{from line 6 to here: standard conjugate gradient}
\LState \raisebox{0pt}[0pt][0pt]{$\phantom{\alpha_{j-1}}\mathllap{\mu_j} = $ Solve $W^TAW\mu_j=W^TAr_j$} \Comment{deflation for following iteration}
\LState \raisebox{0pt}[0pt][0pt]{$\phantom{\alpha_{j-1}}\mathllap{p_j}=\beta_{j-1}p_{j-1}+r_j-W\mu_j$}
\If{$j<\ell$}
\LState \raisebox{0pt}[0pt][0pt]{Store $d_j$, $\alpha_j$, $\beta_j$, $\mu_j$, $p_j$}
\EndIf
\EndWhile
\EndProcedure
\end{algorithmic}
\caption{Deflated Conjugate Gradient method.}
\label{alg:defcg}
\end{algorithm}

\subsection{Approximate Eigenvectors}
\label{sub:approximate_eigenvectors}
One way of obtaining approximate eigenvectors is with the Lanczos algorithm and extract Ritz value/vector pairs. The Lanczos algorithm from which \citet{saad00} derived def-CG, generates a sequence of vectors $\{\vec{p}_j\}$  such that
\begin{equation*}
\vec{p}_{j+1} \perp_A \spa\{W,\vec{p}_0,...,\vec{p}_j\}.
\end{equation*} 
The matrix $A$ can then be transformed into a symmetric and partly tridiagonal matrix $T_{l+k}=Z^TAZ$, with $Z=[W,P_l]\in \Re^{n\times (\ell +k)}$ and $P_\ell=[\vec{p}_0,...,\vec{p}_{\ell-1}]$. The eigendecomposition of $T_{l+k}$ produces pairs $(\theta_j,\vec{u}_j)_{j=1,...,\ell+k}$, of which $\theta_j$ are Ritz values that approximate the eigenvalues of $A$. The corresponding approximate eigenvector is obtained as the Ritz vector $\vec{v}_j=Z \vec{u}_j$. For an orthogonal projection technique, such as deflation, the residual of $(A-\theta I)\vec{v}$ should be orthogonal to $Z$ \cite{chapman97}, leading to the generalized eigenvalue problem
\begin{equation*}
Z\Trans (A-\theta I) Z \vec{u}=0 \q \Rightarrow \q Z\Trans A Z\vec{u}=\theta Z\Trans Z \vec{u}.
\end{equation*}
This can be transformed to a normal eigenvalue problem by multiplying both sides with $(Z\Trans Z)^{-1}$, which exist because the columns in $Z$ are linearly independent, but the symmetric properties would be lost. Computing these matrices generates an additional, non-negligible computational cost, because CG does not generate orthogonal, but $A$-conjugate directions.\footnote{Despite its name, the gradients produced by CG are actually orthogonal, not conjugate. The name arose out of the historical context, because it is a \emph{conjugate} directions method that uses \emph{gradients}.} By instead considering the base $AZ$ for orthogonality, \citet{morgan95} rephrased the problem as a Galerkin method for approximating the eigenvalues of the inverse of $A$  
\begin{equation*}
(AZ)^T(AZ\vec{u} - \theta Z\vec{u})=0.
\end{equation*}
This method is referred to as \textit{harmonic projection}. 
By introducing 
\begin{equation*}
F=(AZ)^TZ, \qquad \qquad G=(AZ)^T(AZ),
\end{equation*}
the problem is conveniently formulated as 
\begin{equation}
G\vec{u}=\theta F\vec{u}.
\label{eq:eig}
\end{equation}
By using properties of the search directions and residuals generated from Algorithm \ref{alg:defcg}, \citet{saad00} explicitly formed $F$ and $G$ from sparse matrices containing the stored quantitities from the first $\ell$ iterations of Algorithm \ref{alg:defcg}. This effectively reduces the computational overhead of finding approximate eigenvectors and makes the method competitive. Once Eq.~\eqref{eq:eig} is solved, one chooses $k$ of the $\ell+k$ Ritz values $\theta$ with corresponding vectors $\vec{u}$ and stores them in $U$. The $k$ approximate eigenvectors for the next system are obtained as $W=ZU$. Preferably the largest or smallest $\theta$ are chosen but this is not required.   

\paragraph{Remark: Connection to First-Order Methods}
\label{sub:connection_to_first_order_methods}

In very high-dimensional machine learning models, second-order optimization methods based on the quadratic model of Eq.~\eqref{eq:optimization-form} are not as popular as first-order methods, like (stochastic) gradient descent and its many flavors. Although we do not further consider this area below, it may be helpful to note that the schemes described here have a weak connection to it, at least in the case of noise-free gradients. That is because the first step of CG is simply gradient descent. The recycling schemes described above can thus be compared conceptually to methods like momentum-based gradient descent \cite{polyak1964some} and other acceleration techniques.

\section{Experiments}
\label{sec:experiments}

We test the utility of def-CG in a classical machine learning benchmark: The infinite MNIST~\cite{loosli06} suite is a tool to automatically create arbitrary size datasets containing images of ``hand-written'' digits, by applying transformations to the classic MNIST set\footnote{\texttt{http://yann.lecun.com/exdb/mnist/}} of actual hand-written digits. We used it to generate a training set $X$ of $36\,551$ images for the digits three and five, each of size $28\times 28$ gray-scale pixels (This means the training set is three times larger than the set of threes and fives in the original MNIST set). By the standards of kernel methods, this is thus a comparably big data set. We consider binary probabilistic classification on this dataset, and follow a setup made popular by \citet{kuss06}. This setting is a good example of the role of linear solvers in machine learning. It involves two nested loops of repeated linear optimization problems. %

This involves computing a (Gaussian) Laplace approximation to the posterior arising from a Gaussian process prior on a latent function $f$ (in our case, $p(f)=\GP(0,k)$, with the Gaussian/RBF kernel $k(\vec{x}_i,\vec{x}_j)=\theta^2 \exp \left(-\nicefrac{(\vec{x}_i-\vec{x}_j)^2}{2\lambda^2}\right)$), and the logistic link function $p(y_i \g f_i) = \sigma(y_i f_i) = 1/(1+e^{-y_if_i})$ as the likelihood (see also \cite[][\textsection 3.7.3, which also outlines the explicit algorithm]{RW-GPML}). 

The outer loop will find the optimal hyperparameters for the kernel and the inner (which is the focus of this study) will find the $\vec{f}$ that maximize
\begin{equation}
\Psi(\vec{f})=\log p(\vec{y}|\vec{f}) + \log p(\vec{f}|X)=\log p(\vec{y}|\vec{f})-\frac{1}{2}\vec{f}^TK^{-1}\vec{f} - \frac{1}{2}\log |K| -\frac{n}{2} \log 2\pi,
\label{eq:obj}
\end{equation}
for a given kernel matrix $K$ with Newton's method.

Newton's method converges to an extremum by evaluating the Jacobian and Hessian of a function at the current location $\vec{x}_n$ and finds a new location by computing $\vec{x}_{n+1}=\vec{x}_n-\operatorname{Hess}_{(\Psi)}^{-1}\operatorname{Jac}_{(\Psi)}$. These iterations involve the solution of a linear system that changes in each iteration.
For Laplace approximation the system to be solved can be made numerically stable by restructuring the computations \cite{kuss06}. In each iteration the target and matrix get a new value 
\begin{align}
\vec{b}^{(i)} &=H^{[i] \frac{1}{2}}K(H^{(i)}\vec{f}_{X}^{(i)} + \nabla \log p(y|\vec{f}_X^{(i)})),\\
A^{(i)} &=I+H^{(i)\frac{1}{2}}K H^{(i) \frac{1}{2}},
\label{eq:linsys}
\end{align}
with $H=-\nabla \nabla \log p(\vec{y}|\vec{f}_X^{(i)})$. This restructuring assures that the eigenvalues $\lambda_i$ of $A$ are contained in $[1,n \max_{ij}(K_{ij})/4]$ and is therefore well-conditioned for most kernels~\cite{RW-GPML}. 

Note how this task fits the setting sub-space recycling methods are designed for: It is difficult to analytically track how an update to $\vec{f}_X^{(i)}$ affects the elements of $(A^{(i)})^{-1}$ and $\vec{b}^{(i)}$, due the non-linear dependence in $\vec{f}_X$. But as the Newton optimizer converges, the iterates change less and less: $|\vec{f}_X^{(i)}-\vec{f}_X^{(i-1)}|<|\vec{f}_X^{(i-1)}-\vec{f}_X^{(i-2)}|$. Thus, $A^{(i)}$ and $\vec{b}^{(i)}$ will change less and less between iterations, and subspace recycling should become increasingly advantageous---up to a point, because of course we have to limit the dimensionality of the deflated space for computational reasons.

Solving the linear system in Eq.~\eqref{eq:linsys} with a Cholesky decomposition is $\mathcal{O}(n^3)$ expensive, where $n$ is the dimension of $A$ i.e.~the size of the training set $X$. Iterative methods, such as CG due to the kernel matrix $K$ being SPD, have been used to speed up the mode-finding of $\hat{\vec{f}}$~\cite{RW-GPML,davies14}. Table \ref{tab:acc} compares the cumulative computational cost and accuracy of the exact Cholesky decomposition, standard CG and deflated CG for each iteration in Newtons method. A reduction of the relative error $\epsilon=|\vec{b}-A\vec{x}_i|/|\vec{b}|$ was used as stopping criterion and was chosen to be $10^{-5}$. 


\setlength{\tabcolsep}{5.4pt} 
\begin{table}
	\caption{Iterative solvers operating on the MNIST classification task. The table shows the progress over Newton iterations. Within each Newton iteration the system in Eq.~\eqref{eq:linsys} needs to solved. Both iterative solvers were set to run until they achieve a relative error of $\epsilon=10^{-5}$. The column labeled $t$ shows cumulative runtimes for each method with the time to extract $W$ included for def-CG.}
	\label{tab:acc}
	\centering
	\begin{tabular}{c|cc|ccc|ccc}
	\toprule
	 & \multicolumn{2}{|c|}{Cholesky} &\multicolumn{3}{|c|}{CG} &\multicolumn{3}{|c}{def-CG($k=8,\,\ell = 12$)}  \\
	\midrule
	   It. & $\log p(\vec{y}|\vec{f})$ & $t\,[\mathrm{s}]$  & $\log p(\vec{y}|\vec{f})$ & rel. error $\delta$ & $t\,[\mathrm{s}]$ & $\log p(\vec{y}|\vec{f})$ & rel. error $\delta$ & $t\,[\mathrm{s}]$\\
	\midrule
	1 & -4926.523 & 426& -4968.760 &$8.573\cdot10^{-3}$ &231 &  -4968.760 & $8.573\cdot10^{-3}$ & 245   \\
	2 & -1915.537 &896 & -1931.348 & $8.254\cdot10^{-3}$&492 &  -1938.585 & $1.203\cdot10^{-3}$ & 436   \\
	3 & -919.124  &1366 & -924.891 & $6.274\cdot10^{-3}$&715 &  -926.668 & $8.208\cdot10^{-3}$ &  617  \\
	4 & -549.182  &1875 & -551.432 & $4.097\cdot10^{-3}$&920 &  -551.796 & $4.760\cdot10^{-3}$ &  790  \\
	5 & -407.058  &2362 & -408.010 & $2.339\cdot10^{-3}$&1088 &  -408.133 & $2.641\cdot10^{-3}$ &  947  \\
	6 & -353.632  &2856 & -354.040 & $1.154\cdot10^{-3}$&1246 &  -354.085 & $1.281\cdot10^{-3}$ &  1101  \\
	7 & -335.575  & 3342& -335.711 & $4.053\cdot10^{-4}$& 1444&  -335.744 & $5.036\cdot10^{-4}$ &  1258  \\
	8 & -331.326  &3815& -331.346 & $6.036\cdot10^{-5}$&1647 &  -331.355 & $8.753\cdot10^{-5}$ &  1418  \\
	9 & -330.997  &4317& -330.998 & $2.309\cdot10^{-6}$&1821 &  -330.996  & $4.018\cdot10^{-6}$ &  1571  \\
	\bottomrule
	\end{tabular}
\end{table}
Table \ref{tab:acc} shows that iterative methods on their own already save computations. Figure~\ref{fig:results} also shows a plot of these results. The increasingly steep downward slope of the CG error in later optimization problems suggests that the systems are getting easier to solve, possibly because the matrix $A^{(i)}$ becomes better conditioned. But the figure also shows the additional advantage of sub-space recycling in def-CG. Eight approximate eigenvectors were used, yielding a saving of at least 12 CG iterations per system ($\sim 25\%$). Enough to outweigh the the computational overhead of finding $W$ and $AW$. Initially, the reduction of iterations for def-CG stagnate (become parallel to CG) over the course of a few Newton iterations, which suggests the recycled subspace fails to reduce the effective condition number further. Either the recycled vectors do not find good approximations to the extreme eigenvalues and the algorithm can be improved, or the difference between successive $A^{(i)}$ is too significant so the propagated information is less useful. Additional experiments (not shown) suggest that both effects play a role, but the former, i.e.~numerical stability, dominates. 
Methods that try to alleviate this problem by estimating the convergence of the approximate eigenvectors exist \cite{gosselet13}, but cause additional computational overhead.

\setlength{\figwidth}{.5\textwidth}
\setlength{\figheight}{0.61803398875\figwidth}

\begin{figure}[t]
    \centering \scriptsize
        \hfill
\begin{tikzpicture}

\definecolor{color1}{RGB}{000,125,122} 
\definecolor{color0}{HTML}{FF9933} 
\definecolor{color2}{RGB}{130,0,0} 

\begin{axis}[
xlabel={Newton iteration},
ylabel={CPU time [s]},
xmin=0.65, xmax=8.35,
ymin=0, ymax=550,
width=\figwidth,
height=\figheight,
tick align=outside,
x grid style={lightgray!92.026143790849673!black},
y grid style={lightgray!92.026143790849673!black},
legend style={draw=white!80.0!black,at={(axis cs:5,300)},anchor=west},
legend entries={{CG},{def-CG(8,12)},{Cholesky}},
legend cell align={left},
mystyle
]
\addplot [semithick, color0, solid, mark=square*, mark size=1, mark options={solid}]
table {%
1 230.4
2 177.648
3 156.018
4 126.346
5 112.917
6 113.662
7 109.538
8 113.281
};
\addplot [semithick, color1, solid, mark=diamond*, mark size=1, mark options={solid}]
table {%
1 243.445
2 130.813
3 127.016
4 105.099
5 94.049
6 99.265
7 86.129
8 99.452
};
\addplot [semithick, color2, solid, mark=*, mark size=1, mark options={solid}]
table {%
1 460.978
2 494.108
3 510.731
4 533.842
5 494.803
6 512.84
7 509.529
8 525.17
};
\end{axis}

\end{tikzpicture}\hfill%
\begin{tikzpicture}

\definecolor{color1}{RGB}{000,125,122} 
\definecolor{color0}{HTML}{FF9933} 

\begin{axis}[
xlabel={Newton Iteration},
ylabel={CG Iterations},
xmin=0.65, xmax=8.35,
ymin=0, ymax=105,
width=\figwidth,
height=\figheight,
tick align=outside,
x grid style={lightgray!92.026143790849673!black},
y grid style={lightgray!92.026143790849673!black},
legend style={draw=white!80.0!black},
legend entries={{CG},{def-CG(8,12)}},
legend cell align={left},
mystyle
]
\addplot [semithick, color0, mark=square*, mark size=1, mark options={solid}]
table {%
1 100
2 83
3 66
4 54
5 50
6 50
7 49
8 48
};
\addplot [semithick, color1, solid, mark=diamond*, mark size=1,mark options={solid}]
table {%
1 100
2 56
3 50
4 43
5 39
6 39
7 37
8 36
};
\end{axis}

\end{tikzpicture}\hfill\null
    \caption{\textbf{Left:} Computational cost (CPU time) per iteration of Newton's method. \textbf{Right:} number of iterations required for CG and def-CG(8,12) to solve a single system (Eq.~\eqref{eq:linsys}) to a relative error of $\epsilon=10^{-5}$. The stopping criterion for the Newton iteration was $\Delta \Psi(\vec{f}) < 1$, thus only requiring the first two terms in Eq.~\eqref{eq:obj} to be computed.}\label{fig:results}
\end{figure}
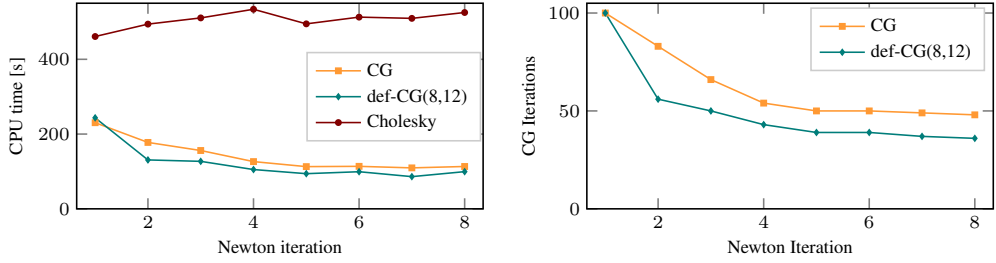
\setlength{\figwidth}{.9\textwidth}

\begin{figure}[b]
    \centering \scriptsize
        \input{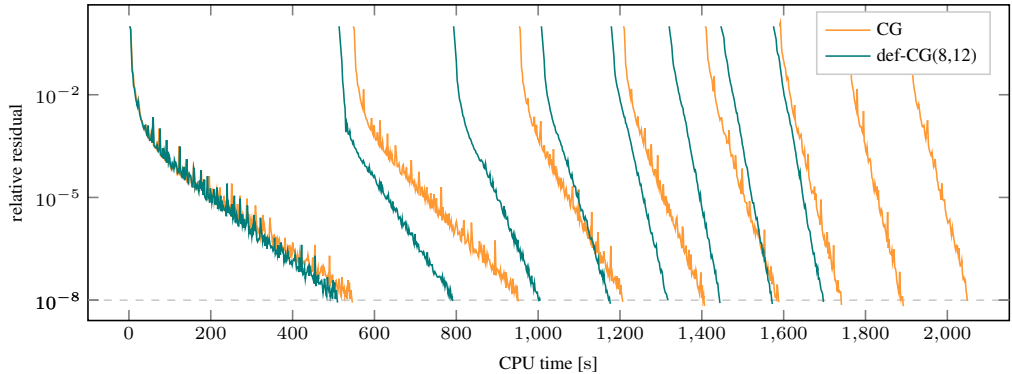}
    \caption{\label{fig:comparison}Relative residual over multiple solutions of Newton's method i.e. the systems described in Eq.~\eqref{eq:linsys} were solved. Each solver had a relative error of $\epsilon=10^{-8}$ as stopping criterion. Similarly to the results in Fig.~\ref{fig:results}, the time required to find a solution that satisfies the error bound becomes faster for each Newton iteration. The deflated method achieves a faster convergence rate as seen by the slope of the relative residual. This confirms that re-using information from a previous system indeed lowers the effective condition number.}

\end{figure}

To better understand the effect of the deflation on individual solutions of Eq.~\eqref{eq:linsys}, Figure~\ref{fig:comparison} compares the convergence of def-CG and standard CG. Each solver was set to run until a relative error of $10^{-8}$ was achieved and the results are shown in Fig.~\ref{fig:comparison}. The figure indicates that the computational savings do not stem from the initial projection onto the A-orthogonal complement of $W$ as one could suspect, but rather from a steeper convergence. This fits with the idea that deflation lowers the effective condition number of deflated system relative to the original problem. 


\setlength{\figwidth}{.99\textwidth}
\setlength{\figheight}{0.41803398875\figwidth}

\setlength{\figwidth}{.9\textwidth}

\subsection{Comparison to Linear-Cost Approximations}

We now investigate the utility of sub-space recycling relative to the linear cost approximation methods of finite error discussed above in Section~\ref{sub:relation_to_linear_cost_methods}, in particular to inducing point methods. These methods assume that the training set elements $\vec{f}_n$ at $X\in \Re^{n\times d}$ are approximately independent of each other when conditioned on a smaller set of values $\vec{f}_m$ at representer points $X_m\in \Re^{m\times d}$. In GPC, this effectively reduces the task to optimizing only the latent variables $\vec{f}_m\in \Re^{m}$ with $m<n$, to maximize the objective in Eq.~\eqref{eq:obj}. The latent variables for the remaining points in the data set are then \emph{induced} by the conditional distribution $p(\vec{f}_{n-m}\g\vec{f}_m)$, with mean $\Exp [\vec{f}_{n-m}\g\vec{f}_m]=K_{(n-m)m} K^{-1}_{mm}\vec{f}_m$. A measure of the performance over the training set is then obtained by evaluating the objective with the inferred latent variables.  

We compared the accuracy of iterative methods to inducing point methods with a randomly selected subset of the training data. Figure \ref{fig:convlin} shows the convergence of the Newton optimizer for subsets $X_m$ of $\log p(\vec{y}\g \vec{f})$ of varying sizes. (each time, error was evaluated on the entire training set $X$). The results confirm the expected picture: The approximate methods can be significantly faster than the iterative solves, but they also incur a significant approximation error. If an accurate solution is required, the iterative solvers can be competitive. For this experiment, the iterative methods have a computational cost comparable to that of the approximate methods running on a (comparably large) subset of between $25\%$ and $50\%$ of the data set. But they also achieve an improvement of about 6 orders of magnitude in precision.

\setlength{\figwidth}{.99\textwidth}
\setlength{\figheight}{0.41803398875\figwidth}
\begin{figure}
    \centering \scriptsize
\begin{tikzpicture}

\definecolor{color0}{HTML}{FF9933} 
\definecolor{color1}{RGB}{000,125,122} 
\definecolor{color2}{RGB}{130,0,0}  
\definecolor{color3}{rgb}{0,0,0}
\definecolor{color4}{rgb}{0.35557281,0.35557281,0.35557281}
\definecolor{color5}{rgb}{0.50838015,0.50838015,0.50838015}
\definecolor{color6}{rgb}{0.65,0.65,0.65}
\definecolor{color7}{rgb}{0.81,0.81,0.81}
\definecolor{color8}{rgb}{0.87,0.87,0.87}




\newcommand{\mSize}{1}
\tikzset{mark size=0.7}

\begin{axis}[
xlabel={CPU time},
ylabel={rel. $\log\, p(\vec{y} | \vec{f})$},
xmin=-242.75775, xmax=5407.70275,
ymin=6.61209485199062e-07, ymax=32393.1774573419,
ymode=log,
width=\figwidth,
height=\figheight,
tick align=outside,
x grid style={lightgray!92.026143790849673!black},
y grid style={lightgray!92.026143790849673!black},
legend style={draw=white!80.0!black},
legend entries={{CG},{def-CG(8,12)},{100\%},{95\%},{75\%},{50\%},{25\%},{10\%},{5\%}},
legend cell align={left},
mystyle
]
\addplot [semithick, color0, mark=square*, mark size=0.9, mark options={solid}]
table {%
234.234 14.0115848254505
472.974 4.9002182217254
662.053 1.81563625976223
842.217 0.675879540476442
1008.405 0.237160064955664
1179.182 0.0715239263432982
1353.027 0.0149403326334234
1514.169 0.00117208477469438
1687.844 8.35118355262598e-06
};
\addplot [semithick, color1, mark=diamond*, mark size=1.1, mark options={solid}]
table {%
228.019 14.0115848245442
412.137 4.83608200728108
591.732 1.7946146588317
766.293 0.665719763138416
913.287 0.232615263070439
1068.877 0.0695473617426246
1214.731 0.0141987582894002
1359.045 0.00105346002205464
1500.274 2.02419976136714e-06
};
\addplot [semithick, color2, mark=*, mark options={solid}]
table {%
425.113 13.8839801205456
891.749 4.78720826598589
1401.94 1.77685161407272
1918.046 0.659185184066225
2418.924 0.229801054396592
2912.795 0.0683907611897461
3404.559 0.0138370670251816
3879.709 0.00100001510596842
4382.189 6.04238734726975e-06
4864.5 0
};
\addplot [semithick, color3, mark=*, mark options={solid}]
table {%
389.218 2464.93800208462
789.425 917.61264067433
1185.997 374.927944530884
1589.675 152.579691536126
2003.268 51.206969893805
2408.886 11.1672653665463
2810.647 0.816643755947975
3221.507 0.0794694783909122
3634.548 0.0761975256423813
4057.781 0.0762277375791175
};
\addplot [semithick, color4, mark=*, mark options={solid}]
table {%
174.27 10581.305560507
368.783 3939.59855284823
592.393 1595.68232752761
767.575 625.125143884349
943.809 195.497089079895
1115.593 38.1031616791794
1298.459 2.35554917748002
1478.111 0.509282617562199
1671.194 0.501107267481382
1848.437 0.501128415837097
};
\addplot [semithick, color5, mark=*, mark options={solid}]
table {%
54.124 10257.2859076421
109.123 3856.45010045469
164.144 1588.19623257149
219.005 628.721297904802
273.996 192.882309400444
338.003 32.1224308524298
395.248 2.14376350095923
450.256 1.28404356561277
505.258 1.28316741944742
560.291 1.28316741944742
};
\addplot [semithick, color6, mark=*, mark options={solid}]
table {%
11.118 6991.35015936797
22.863 2656.30398948625
34.562 1092.69346062629
46.299 417.827746642699
56.835 117.635631958187
68.639 11.3683952929803
80.395 2.81854710796236
90.92 2.74849469025212
101.537 2.74839499086089
};
\addplot [semithick, color7, mark=*, mark options={solid}]
table {%
1.695 3129.07345428179
3.986 1157.92056073355
6.263 423.819701203946
8.538 118.144322421789
10.813 19.094992371486
13.085 5.72625266242693
15.357 5.54286922763184
17.634 5.54517137721114
};
\addplot [semithick, color8, mark=*, mark options={solid}]
table {%
0.445 3454.21567395278
1.025 1227.06122751099
1.53 441.22038701491
2.046 112.616519887007
2.547 14.1225094034653
3.074 8.22942944757474
3.579 8.17629571443677
4.09 8.17619601504554
};%
\end{axis}
\end{tikzpicture}
   \caption{\label{fig:convlin}Comparison of accuracy between the iterative methods CG, def-CG and differently sized subsets of data measured as the relative error of $\log p(\vec{y}\g \vec{f})$ to the ``exact'' (up to machine precision) value achieved by a direct (Cholesky) solver on the full data set. Each dot represents the approximated $\log p(\vec{y}\g \vec{f})$ for the full training set after each iteration of Newton's method. The CPU time refers to the cumulative time spent solving the linear systems in Eq.~\eqref{eq:linsys}, which is the computationally most expensive part of each Newton iteration.}
\end{figure}
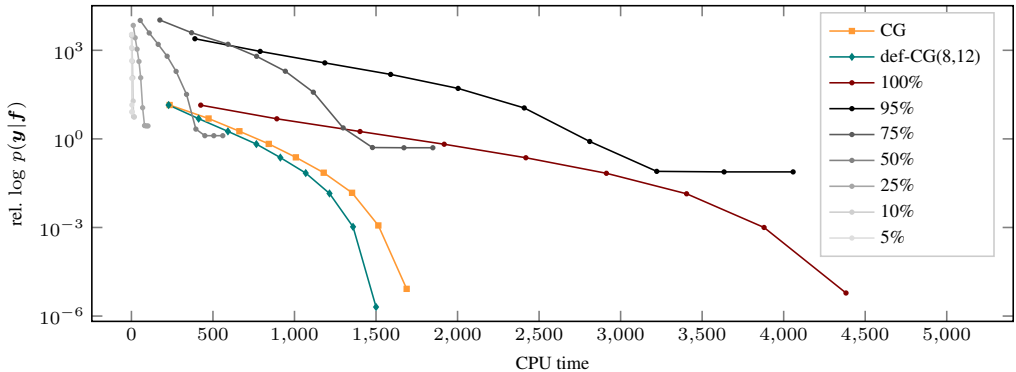
\setlength{\figwidth}{.9\textwidth}



\section{Conclusion} 
\label{sec:conclusion}

We have investigated the use of Krylov sub-space recycling methods for a realistic example application in machine learning. ML problems often involve outer loops of (hyper-) parameter optimization, which produce a sequence of interrelated linear, symmetric positive definite (aka.~least-squares) optimization problems. Subspace-recycling methods allow for iterative linear solvers to share information across this sequence, so that later instances become progressively cheaper to solve. Our experiments suggests that doing so can lead to a useful reduction in computational cost. While non-trivial, sub-space recycling methods can be implemented and used with moderate coding overhead. 

We also compared empirically to the popular option of fixing a low-rank sub-space a priori, in the form of spectral or inducing point methods. The overarching intuition here is that these methods can achieve much lower computational cost when they use a low-dimensional basis. But in exchange they also incur a significant computational error. In applications where computational \emph{precision} is at least as important as computational \emph{cost}, sub-space recycling iterative solvers provide a reliable answer of high quality. While their run-time scales quadratically with data-set size, they are certainly scalable, at acceptable run-times, to data-sets containing $\sim 10^5$ to $\sim 10^6$ data points.

\small
\bibliographystyle{abbrvnat}
\bibliography{bibfile}

\end{document}